\PassOptionsToPackage{table}{xcolor}
\documentclass[10pt,twocolumn,letterpaper]{article}

\usepackage{iccv}              

\definecolor{iccvblue}{rgb}{0.21,0.49,0.74}
\usepackage[pagebackref,breaklinks,colorlinks,allcolors=iccvblue]{hyperref}
\usepackage{times}
\usepackage{url}
\usepackage{booktabs}
\usepackage{multirow, multicol}
\usepackage{makecell}
\usepackage{subcaption}
\usepackage{graphicx}
\usepackage{bbding}
\usepackage{pifont}
\usepackage{alltt}
\usepackage{algorithm}
\usepackage{algorithmic}
\usepackage{wrapfig, lipsum}
\usepackage{caption}
\definecolor{sblue}{HTML}{A6CAEC}

\definecolor{codegreen}{rgb}{0.0, 0.6, 0.0}
\definecolor{codered}{rgb}{0.6, 0.0, 0.0}
\definecolor{codeblue}{rgb}{0.2, 0.2, 1.0}
\definecolor{lightcyan}{rgb}{0.83, 0.94, 0.98}

\def\paperID{4102} 
\def\confName{ICCV}
\def\confYear{2025}

\title{Cross-Architecture Distillation Made Simple with Redundancy Suppression}

\author{Weijia Zhang ~~~~~~~~ Yuehao Liu ~~~~~~~~ Wu Ran ~~~~~~~~ Chao Ma\thanks{Corresponding author}\\
Shanghai Jiao Tong University \\
{\tt\small \{weijia.zhang, yuehao.liu, bonjourlemonde, chaoma\}@sjtu.edu.cn}
}

\begin{document}
\maketitle
\begin{abstract}
We describe a simple method for cross-architecture knowledge distillation, where the knowledge transfer is cast into a redundant information suppression formulation.
Existing methods introduce sophisticated modules, architecture-tailored designs, and excessive parameters, which impair their efficiency and applicability.
We propose to extract the architecture-agnostic knowledge in heterogeneous representations by reducing the redundant architecture-exclusive information. To this end, we present a simple redundancy suppression distillation (RSD) loss, which comprises cross-architecture invariance maximisation and feature decorrelation objectives.
To prevent the student from entirely losing its architecture-specific capabilities, we further design a lightweight module that decouples the RSD objective from the student's internal representations. 
Our method is devoid of the architecture-specific designs and complex operations in the pioneering method of OFA. It outperforms OFA on CIFAR-100 and ImageNet-1k benchmarks with only a fraction of their parameter overhead, which highlights its potential as a simple and strong baseline to the cross-architecture distillation community.
\end{abstract}    
\section{Introduction}
\label{sec:intro}

Knowledge distillation (KD) aims to transfer the privileged capability of a pre-trained teacher model to a usually less capable student to improve its performance.
Since its introduction by ~\citet{kd} in 2015, KD has witnessed significant advancements in methodology and performance over a decade. Existing KD methods extract and transfer different kinds of knowledge, such as network outputs~\cite{kd, dkd, lskd, tinyvit}, intermediate representations~\cite{fitnets, crd, reviewkd, vitkd}, or higher-order correlations amongst them~\cite{rkd, cckd, manifold}, and have demonstrated widespread success across different computer vision applications~\cite{cirkd, fgfi, fgd, odm3d, unidistill, discofos, ekd, hfgi3d,  kddlgan}. Yet, previous works mostly considered distillation between models of the same architectural type (\textit{e.g.}, CNNs~\cite{resnet}).

Recently, with the rise of novel vision architectures such as ViTs~\cite{vit, swin} and MLPs~\cite{mlpmixer, resmlp}, it becomes increasingly relevant for the community to study knowledge distillation across distinct architectures, a task known as cross-architecture knowledge distillation (CAKD). 
Distilling across heterogeneous architectures finds broad practical relevance, particularly given that the best-performing models today are often not the most deployment-friendly ones. 
Nonetheless, contrary to a thriving literature on generic KD, cross-architecture KD remains a less explored setting, with OFA~\cite{ofa} being a pioneering method.  

\begin{figure}[t]
\centering
\includegraphics[width=0.98\linewidth]{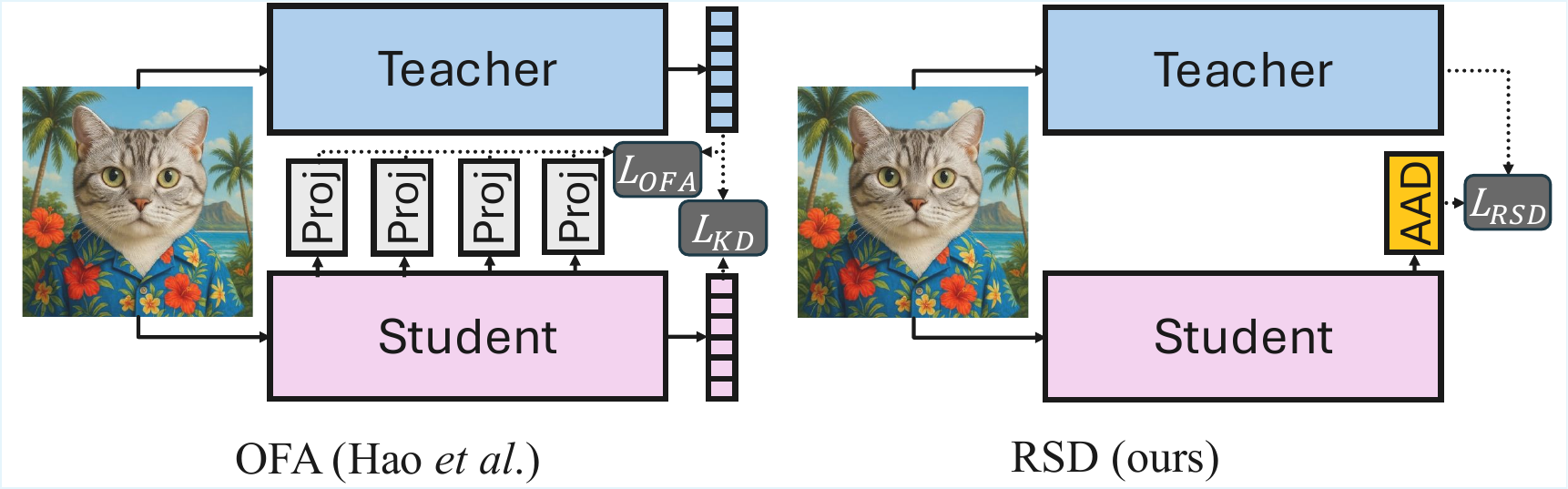} \\
\includegraphics[width=\linewidth]{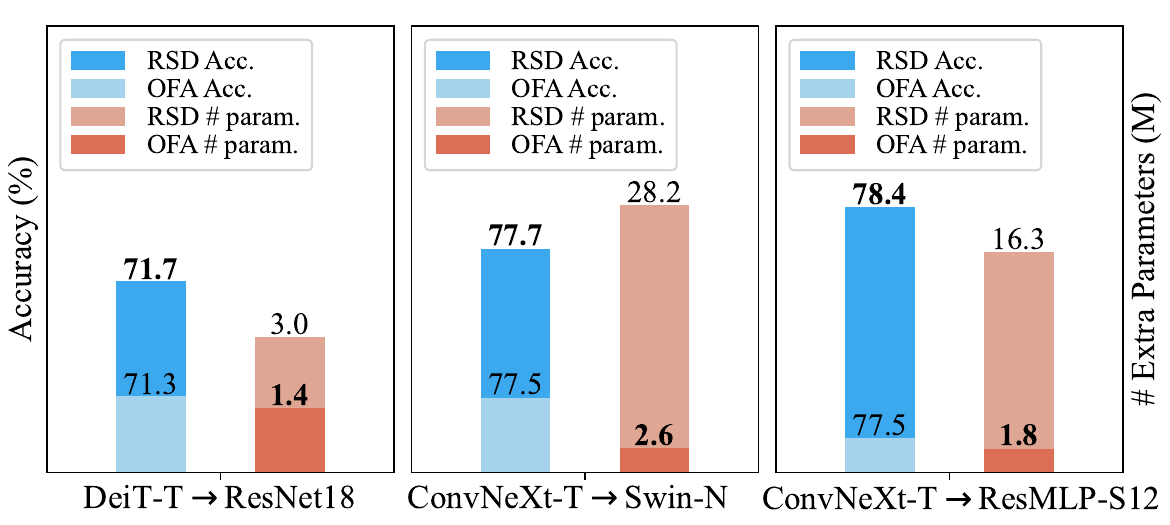}
\vspace{-0.6cm}
\caption{\textbf{A comparison of RSD and OFA.} Compared to the pioneering method of OFA, RSD is simpler in design, stronger in performance, and lower in complexity.}
\label{fig:teaser}
\vspace{-0.3cm}
\end{figure}

Compared to generic distillation, CAKD is more complicated, which is fundamentally due to the distinct properties of  representations from different architectures.  
First, heterogeneous features have \textit{different dimensionalities}, which requires additional operations to reconcile. 
They also possess \textit{distinct, even conflicting patterns and characteristics}~\cite{texturebias, raghu21, park22}. Forcing the student to blindly absorb heterogeneous information leads to degraded performance~\cite{cakd, park22}. 
As a result, generic KD methods show limited effectiveness or even collapse when directly applied to CAKD, as shown in Tables~\ref{tab:cifar} and~\ref{tab:imagenet}.

As a seminal method, OFA has to make complex, architecture-tailored designs to unify heterogeneous feature dimensionality, such as depth-wise separable convolution~\cite{dsconv} for CNN features, and token operations and attention mechanisms~\cite{vit} for ViT and MLP features.
To circumvent the representation heterogeneity, OFA projects intermediate features into the logit space, which is considered architecture-agnostic. However, useful fine-grained knowledge is lost in this process. 
OFA attaches separate modules to each of the network stages to retain more information, which amounts to quadrupled parameter complexity. All these designs render OFA complicated and inefficient.
For instance, when performing ConvNeXt-T-to-Swin-N distillation, the projectors introduced by OFA are about 3 times larger than the student itself in terms of the number of parameters (\textit{i.e.,} 28.2M v.s. 9.6M), shown in Figure~\ref{fig:efficiency}, making it costly and less practical in real-world scenarios.

In this paper, we present a new method for cross-architecture distillation. Rather than projecting features into the logit space, we learn to explicitly extract the architecture-agnostic knowledge. We achieve this via the proposed redundancy suppression distillation (RSD) objective, rooted in information maximisation~\cite{tishby2000, tishby2015} and feature decorrelation~\cite{barlow, deco1997} principles. This allows the student to grasp the underlying architecture-independent knowledge, without being hampered by redundant, distracting patterns specific to the teacher architecture. We also insert a lightweight MLP module to decouple the student's internal representations from the direct dictation of the RSD objective, which enables the student to preserve some beneficial capabilities unique to its own architecture. A high-level comparison of our method to OFA is made in Figure~\ref{fig:teaser}.

The advantages of our method are manifold. First and foremost, it only involves a simple loss function and a lightweight MLP-based gadget. Since RSD is applied to the penultimate-layer embeddings, our method is devoid of the complex and tailored adaptation modules for unifying heterogeneous features in OFA. The embeddings also offer a richer amount of knowledge than the logit space of OFA. 
RSD does not rely on memory banks~\cite{crd}, assistant networks~\cite{takd}, teacher weight reuse~\cite{simkd, fcfd}, attention mechanisms~\cite{tat, cakd}, adversarial training~\cite{cakd},
or asymmetric transformation~\cite{cakd, crld}. 
Besides, it does not require access to the full-stage intermediate features~\cite{fitnets, ofa}, and is shown to also work reasonably well on model logits, adding to its value in practical black-box distillation scenarios where security and privacy concerns are relevant.

In summary, this paper makes several significant contributions to the knowledge distillation literature. We provide a redundancy suppression perspective to cross-architecture distillation, which motivates us to explicitly get rid of the redundant, teacher-specific information during the distillation process. 
We describe a simple redundancy suppression criterion based on information maximisation and feature decorrelation objectives. We show that moderate preservation of the student's inherent characteristics further benefits cross-architecture distillation.
We report substantial performance improvements over OFA~\cite{ofa} using a fraction of its parameter costs. RSD may potentially be a strong and efficient baseline for the community and inspire future endeavours to cross-architecture knowledge distillation.
\section{Related Work}
\label{sec:related_work}

\subsection{Vision architectures}
\paragraph{Convolutional Neural Networks (CNNs)} have been the cornerstone behind modern computer vision. They rely on convolution operations to capture local spatial patterns within images and pooling to reduce dimensionality. The shared weights of convolution equip CNNs with desirable properties such as inductive biases and translational invariance. CNNs are popularised by early pioneers such as AlexNet~\cite{imagenet} and VGGNet~\cite{vgg}. ResNet~\cite{resnet} introduces residual connections to enable much deeper and more capable CNNs. Efficient CNNs~\cite{shufflenet, mobilenetv2} are also engineered to facilitate their real-world applications. Recent models such as ConvNeXt~\cite{convnext} also incorporate design principles from Vision Transformers for better efficiency and performance.

\paragraph{Vision Transformers (ViTs)} are an adaptation of Transformers~\cite{transformer} in natural language processing (NLP) to the vision domain. ViTs~\cite{vit} treat images as sequences of patches and process them using self-attention mechanisms, which allows them to flexibly capture long-range dependencies and global context within the image. Following ViT's initial success, Swin Transformer~\cite{swin} introduces hierarchical window-based attention for improved training speed and performance, whereas DeiT~\cite{deit} enables effective ViT training on smaller datasets via distillation.

\paragraph{Multi-Layer Perceptrons (MLPs)}  recently emerged as a competitive alternative to CNNs and ViTs in vision tasks. They convert images into flattened vectors and process them via fully-connected (FC) layers. MLP-Mixer~\cite{mlpmixer} and ResMLP~\cite{resmlp} introduced novel mixing operations, such as token-mixing and channel-mixing, to capture spatial dependencies without convolutions or self-attention. 
The simplicity, scalability, and efficiency of MLPs make them an appealing alternative to more complex models.

\begin{figure*}[t] \centering
\includegraphics[width=0.88\textwidth]{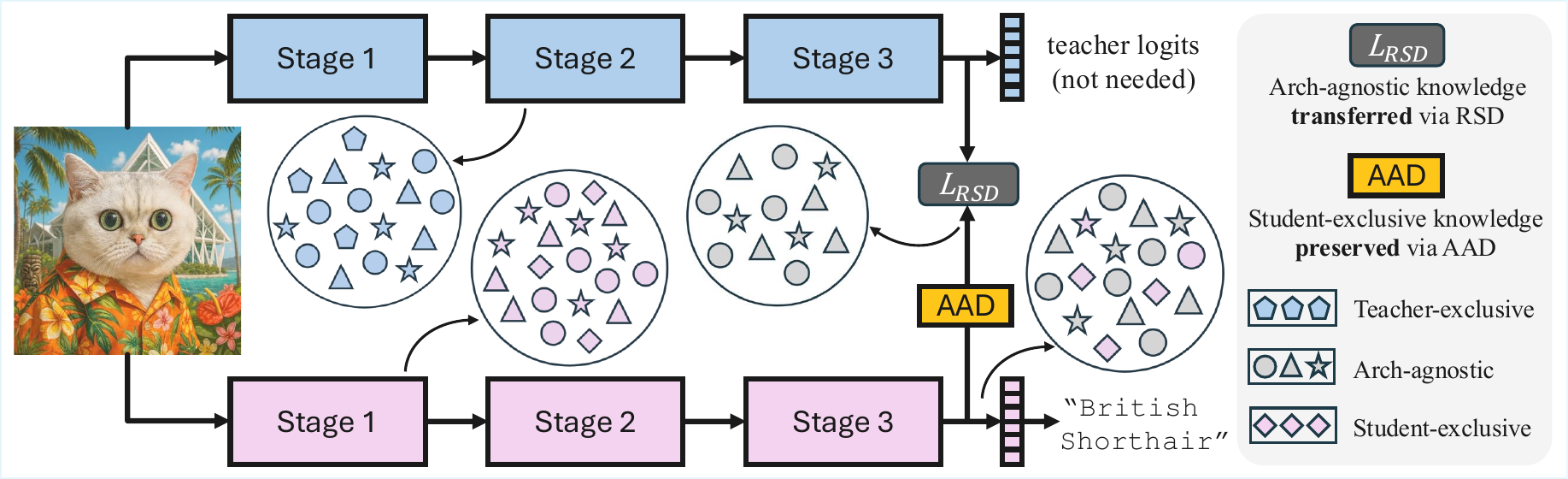} 
\caption{\textbf{A schematic diagram of RSD for cross-architecture knowledge distillation.} RSD employs a redundancy suppression objective (\textit{i.e.}, RSD loss) to extract and transfer architecture-agnostic knowledge that is common to both teacher and student architectures. It uses a lightweight Architecture-Agnostic Knowledge Decoupler (AAD) module to align student representation dimension to the teacher's, while decoupling the RSD dictation to allow the preservation of useful student-exclusive knowledge within the student.}
\label{fig:method} \vspace{-0.1cm}
\end{figure*}

\subsection{Knowledge distillation}
\paragraph{Generic KD.} First proposed in~\cite{kd} as a model compression technique, knowledge distillation (KD) transfers the privileged capability of a usually more capable pretrained teacher model to a lightweight student model. KD methods can be broadly categorised according to what sort of knowledge is extracted and transferred to the student. A group of methods~\cite{kd, dkd, nkd, ctkd, lskd, ttm, sdd, e2kd, crld, tinyvit} match the student's prediction logits to the teacher's, and are called logits-based methods. By contrast, feature-based methods~\cite{fitnets, ofd, crd, catkd, reviewkd, semckd, tat, ndkd, vitkd} let the student learn to mimic the intermediate features of the teacher. An orthogonal concept to the above methods is relational KD, where the student learns to replicate the higher-order correlations constructed from the teacher's features~\cite{rkd, cckd, spkd, ickd, manifold} or logits~\cite{dist, vrm}. All these methods consider homogeneous architecture distillation, such as CNN to CNN~\cite{kd, fitnets, crd, reviewkd, dkd, tat, nkd, lskd, ttm, sdd, crld} or ViT to ViT~\cite{vitkd, manifold, tinyvit}.

\paragraph{Cross-architecture KD.}
While a few works~\cite{deit, tinytransformer, dearkd} made early attempts at unidirectional CNN-to-ViT distillation under basic settings, it is only until recently that the pioneering work of OFA~\cite{ofa} proposed a universal paradigm alongside rigorous evaluation settings for diverse cross-architecture distillation between CNNs, ViTs, and MLPs. 
OFA employs a set of projectors to transform intermediate features into the architecture-agnostic logit space, each using a target-knowledge enhancement mechanism.  
However, OFA introduces significant training overheads due to its complicated designs and mechanisms. In this work, we present a much simpler orthogonal approach that is able to outperform OFA with a fraction of its parameters.

\subsection{Redundancy in representation learning}
Previous research discovers that redundancy exists in the representations learnt by deep learning models, which motivates a group of works to mitigate the spatial~\cite{gfnet, octconv, scconv} and channel~\cite{slimconv} redundancy in CNN features through network or operator redesigns.
Meanwhile, redundancy reduction is the core principle in early unsupervised feature learning theories~\cite{barlow, atick1990, deco1997}, which have also helped alleviate  informational collapse in recent self-supervised learning algorithms~\cite{vicreg, duality, barlowtwins, disperseloss}.
An orthogonal body of research is along domain generalisation~\cite{dica, dat, csd, difex}. These works aim to learn domain-invariant representations for generalisation to unseen domains, which may be viewed as suppressing the redundant domain-specific information. Yet, these works substantially differ from ours in context and methodology.
In this work, we make initial efforts to improve cross-architecture distillation from a representation redundancy perspective.
\section{Method}

\subsection{Preliminaries} \label{sec:prelim}
\paragraph{Generic KD.} In knowledge distillation (KD), we are given a pre-trained teacher model and a to-be-trained student model. During distillation, the student is simultaneously supervised by the ground-truth labels and distillation signals from the teacher.
For logit KD, the student is trained to mimic the output predictions given by the teacher, often via the Kullback–Leibler divergence (KLD) loss, and the optimisation objective is given by:
\begin{align}
    \mathcal{L}_{\text{logit-kd}} = 
    \text{CE}(\mathbf{z}^s, y) + \lambda \text{KLD}(\mathbf{z}^s, \mathbf{z}^t),
\end{align}
where $\mathbf{z}^s$ and $\mathbf{z}^t$ are the student and teacher logits, respectively, $y$ is the ground-truth label, and CE($\cdot$) is the cross-entropy loss. Whereas for feature KD, the student learns to match its intermediate features to the teacher's from the corresponding layers. The objective is therefore:
\begin{align}
    \mathcal{L}_{\text{feature-kd}} = 
    \text{CE}(\mathbf{z}^s, y) + \lambda \text{MSE}(\psi(\mathbf{f}^s), \mathbf{f}^t),
\end{align}
where the mean-square error (MSE) loss is often used to drive student features $\mathbf{f}^s$ towards teacher features $\mathbf{f}^t$. A convolutional adaptor network $\psi(\cdot)$ is often used to match the student feature dimension to the teacher's~\cite{fitnets}.

\paragraph{OFA.} 
As the first universal method for cross-architecture distillation between CNNs, ViTs, and MLPs, OFA~\cite{ofa} projects intermediate features of the student into a common logit space to be supervised by the teacher logits. Its training objective is given by:
\begin{align}
    \mathcal{L}_{\text{ofa}} = 
    \text{CE}(\mathbf{z}^s, y) &+ \lambda_1 \sum_{i \in S} \text{OFA}(\psi_{\text{ofa},i}(\mathbf{f}^s_i), \mathbf{z}^t) \nonumber \\
    &+ \lambda_2 \text{OFA}(\mathbf{z}^s ,\mathbf{z}^t),
    \label{eq:ofa}
\end{align}
where OFA($\cdot$) is a variant of the KLD loss, and $S$ is the set of stages whose features are used for projection ($|S| = 4$). Each stage $i$ requires a separate projection network $\psi_{\text{ofa},i}$. Each  $\psi_{\text{ofa}}$ is much bulkier than the original $\psi$ in ~\cite{fitnets}, involving complex arch-specific operations such as depthwise separable convolution~\cite{dsconv} or ViT blocks~\cite{vit}.
All these designs in OFA amount to its excessive complexity.

\begin{algorithm}[t]
\caption{PyTorch-style pseudocode for RSD Loss} \label{alg:rrd}
\begin{algorithmic}[1] \footnotesize
\STATE \textcolor{codegreen}{\# feat\_t: teacher features, feat\_s: student features after AAD}
\STATE \textcolor{codegreen}{\# B: batch size, D: feature dimension}

\STATE $\texttt{feat\_t = Normalize(feat\_t)}$ \quad \textcolor{codegreen}{\# [B,D]}
\STATE $\texttt{feat\_s = Normalize(feat\_s)}$ \quad \textcolor{codegreen}{\# [B,D]}

\STATE \texttt{rcc = torch.mm(feat\_t.T, feat\_s)} \quad \textcolor{codegreen}{\# [D,D]}

\STATE $\texttt{idt = \text{torch.eye}(D)}$ \quad \textcolor{codegreen}{\# [D,D]}

\STATE $\texttt{loss\_rsd = \text{torch.mse}(rcc, idt)}$ \quad \textcolor{codegreen}{\# [D,D]}

\STATE \texttt{loss\_rsd[(1 - idt).bool()] *= $\kappa$} \quad \textcolor{codegreen}{\# [D,D]} 
\STATE \texttt{\textcolor{codeblue}{return} loss\_rsd.mean()} 
\end{algorithmic}
\end{algorithm}

\subsection{Redundancy Suppression Distillation}

\paragraph{A redundancy suppression perspective to CAKD.}
In cross-architecture knowledge distillation, the teacher and student representations bear distinct, even conflicting characteristics due to their heterogeneous nature, as have been demonstrated in prior works~\cite{texturebias, raghu21, cmt, park22}. Meanwhile, when fed with the same training data, the representations produced by the teacher and the student, albeit very different, also encode the same semantic information corresponding to the very same input data. We expect such information to be fundamentally \textit{architecture-independent} but only input-dependent (denoted by uncoloured circles, triangles, and stars in Figure~\ref{fig:method}). Yet, this underlying \textit{architecture-agnostic knowledge} is blended with arch-specific patterns (navy and pink colours) or knowledge (pentagons and diamonds) in respective representations. Prior research has shown that blindly dictating these arch-specific patterns over heterogeneous architectures leads to incompatibility and degradation~\cite{cakd, ofa, scalekd, park22}. Intuitively, we would like the student to grasp the arch-independent knowledge, such that it will not be distracted by any teacher-specific patterns when generalising to different heterogeneous teachers. Hence, give a pair of heterogeneous features, we may extract the commonality between them by suppressing the redundant architecture-specific or irrelevant information.

\paragraph{Extracting common knowledge between teacher and student.}
We leverage the correlations between feature units of heterogeneous representations to learn the architecture-agnostic representation, inspired by classic theories on unsupervised feature extraction~\cite{barlow, atick1990, deco1997} and Information Bottleneck (IB) principle~\cite{tishby2000, tishby2015}.
Specifically, we construct matrix $\mathbf{P}\in \mathbb{R}^{D \times D}$ of Pearson correlations between pairs of feature units of the teacher and student representations, \textit{i.e.}, $\mathbf{z}^t, \mathbf{z}^s \in \mathbb{R}^{B \times D}$:
\begin{align}
    \mathbf{P}_{ij} &= \frac{\sum_{k=1}^{B} (\mathbf{z}^t_{ki} - \bar{\mathbf{z}^t_i})(\mathbf{z}^s_{kj} - \bar{\mathbf{z}^s_j})}{\sqrt{\sum_{k=1}^{B} (\mathbf{z}^t_{ki} - \bar{\mathbf{z}^t_i})^2 \sum_{k=1}^{B} (\mathbf{z}^s_{kj} - \bar{\mathbf{z}^s_j})^2}}
    \nonumber \\ 
    & \quad \text{for} \quad i, j \in \{1, \dots, D\},
\end{align}
where $B$ is the batch size and $D$ is the embedding dimension. Essentially, $\mathbf{P}$ captures the similarity in batch-wise activation patterns between pairs of feature dimensions of the teacher and the student.
Each diagonal element in $\mathbf{P}$ represents the correlation between the same feature dimension of teacher and student embeddings. To extract the ``commonality'' between the pair of heterogeneous representations, we learn $\mathbf{z}^s$ that leads to maximised invariance between itself and $\mathbf{z}^t$ \textit{for the same feature unit}~\cite{barlow, deco1997, barlowtwins, disperseloss}. This translates to driving the diagonal elements of $\mathbf{P}$ to 1. Hence, the optimisation target for our task can be built via a target matrix $\mathbf{T}\in \mathbb{R}^{D \times D}$, whereby:
\begin{equation}
\mathbf{T}_{ij} = 1 \quad \text{if } i = j. \label{eq:pearson}
\end{equation}

\begin{table*}[t]
  \centering
  \small
  \renewcommand\tabcolsep{3.5pt}
  \renewcommand{\arraystretch}{0.96}
  \resizebox{0.79\textwidth}{!}{
  \begin{tabular}{cc cc cccc cccc>{\columncolor[gray]{0.85}}c}
    \toprule
    \multirow{2}{*}{Teacher} & \multirow{2}{*}{Student} & \multicolumn{2}{c}{From Scratch} & \multicolumn{4}{c}{Logits-based} & \multicolumn{5}{c}{Feature-based} \\
    \cmidrule(lr){3-4} \cmidrule(lr){5-8} \cmidrule(lr){9-13} & & T. & S. & KD & DKD & DIST & OFA & FitNets & CC & RKD & CRD & \textbf{RSD} \\
    \midrule
    \multicolumn{2}{c}{\emph{CNN-based students}} \\
    \midrule
    Swin-T & ResNet18  & 89.26 & 74.01 & 78.74     & 80.26  & 77.75 & \underline{80.54} & 78.87 & 74.19 & 74.11 & 77.63 & \textbf{83.92} \\ 
    ViT-S & ResNet18 & 92.44 & 74.01 & 77.26 & 78.10 & 76.49 & \underline{80.15} & 77.71 & 74.26 & 73.72  & 76.60 & \textbf{81.50} \\ 
    Mixer-B/16 & ResNet18 & 87.62 & 74.01 & 77.79 & 78.67 & 76.36 & \underline{79.39} & 77.15 & 74.26 & 73.75 & 76.42 & \textbf{81.85} \\ 
    Swin-T & MobileNetV2 & 89.26 & 73.68 & 74.68 & 71.07 & 72.89  & \underline{80.98} & 74.28 & 71.19 & 69.00 & 79.80 & \textbf{83.68} \\ 
    ViT-S & MobileNetV2 & 92.44 & 73.68 & 72.77 & 69.80 & 72.54 & \underline{78.45} & 73.54 & 70.67 & 68.46 & 78.14 & \textbf{81.68} \\ 
    Mixer-B/16 & MobileNetV2 & 87.62 & 73.68 & 73.33 & 70.20 & 73.26 & \underline{78.78} & 73.78 & 70.73 & 68.95 & 78.15 & \textbf{81.74} \\ 
    \midrule
    \multicolumn{2}{c}{\emph{Transformer-based students}} \\ \midrule
    ConvNeXt-T & DeiT-T & 88.42 & 68.00 & 72.99 & 74.60 & 73.55 & \underline{75.76} & 60.78  & 68.01 & 69.79 & 65.94 &  \textbf{82.46} \\ 
    Mixer-B/16 & DeiT-T & 87.62 & 68.00 & 71.36 & 73.44 & 71.67 & \underline{73.90} & 71.05  & 68.13 & 69.89 & 65.35 & \textbf{78.50} \\ 
    ConvNeXt-T & Swin-P & 88.42 & 72.63 & 76.44 & 76.80 & 76.41 & \underline{78.32} & 24.06 & 72.63 & 71.73 & 67.09 & \textbf{82.21} \\ 
    Mixer-B/16 & Swin-P & 87.62 & 72.63  & 75.93 & 76.39 & 75.85 & \underline{78.93} & 75.20 & 73.32 & 70.82 & 67.03 & \textbf{81.28} \\ 
    \midrule
    \multicolumn{2}{c}{\emph{MLP-based students}}  \\
    \midrule
    ConvNeXt-T & ResMLP-S12 & 88.42 & 66.56 & 72.25 & 73.22 & 71.93 & \underline{81.22} & 45.47 & 67.70 & 65.82 & 63.35 & \textbf{84.21} \\ 
    Swin-T & ResMLP-S12 & 89.26 & 66.56 & 71.89 & 72.82 & 11.05 & \underline{80.63} & 63.12 & 68.37 & 64.66 & 61.72 & \textbf{82.67} \\ 
        \midrule
    \textit{Average Gain} & - & - & - & \textit{+3.17} & \textit{+3.16} & \textit{-2.31} & \textit{+7.47} & \textit{-5.20} & \textit{-0.33} & \textit{-1.40} & \textit{-0.02}  & \textbf{\textit{+10.69}} \\ 
    \bottomrule
  \end{tabular}} \vspace{-0.1cm}
   \caption{\textbf{Cross-architecture distillation results on CIFAR-100.} Best results are in bold and second best results  underlined.}
  \label{tab:cifar}
  \vspace{-0.1cm}
\end{table*}

According to early unsupervised feature learning theories, the minimisation of redundancy in features corresponds to extracting statistically independent features~\cite{barlow, atick1990, deco1997}. This motivates us to further suppress the redundancy in learnt representation $\mathbf{z}^s$ by decorrelating its feature units. 
For the cross-architecture distillation problem, we revamp this approach and instead minimise the mutual information between each student feature unit and every other teacher feature unit. This makes the redundancy suppressed within the heterogeneous representation space, which implicitly facilitates the extraction of architecture-agnostic information. Formally, this simply means forcing all off-diagonal elements in $\mathbf{P}$ to zero by:
\begin{equation}
\mathbf{T}_{ij} = 0 \quad \text{if } i \neq j,
\end{equation}
which suggests $\mathbf{T}$ is an identity matrix, and we may use any distance measure $d(\cdot)$ to compute a loss between $\mathbf{P}$ and $\mathbf{T}$.
We empirically find that this decorrelation objective further improves the performance of our method most of the time. Note that in Equation~\ref{eq:pearson} we have assumed identical teacher and student embedding dimensionality $D$, which is not necessarily the case. Our decoupler module will make the alignment for arbitrary teacher and student embeddings, which is described next. 

\paragraph{Retaining exclusive knowledge of the student.}
Ideally, the architecture-agnostic representation learnt through the described redundancy suppression objective would suppress any teacher-exclusive or student-exclusive knowledge.
Since different architectures have their unique~\cite{raghu21, cmt}, sometimes opposite~\cite{park22} characteristics and behaviours, we expect that by moderately preserving the student's unique representation patterns, it may be able to retain some of its exclusive advantages. For instance, a CNN student may exhibit prominent activations to local flurry patterns~\cite{texturebias} in a ``British shorthair" cat image owing to the locality properties of convolution operations. Such ability is largely architecture-exclusive. It is not possessed by an inductive-bias-deficient ViT teacher~\cite{raghu21, convit}, and is therefore absent in their common knowledge. In this case, it would be beneficial to let the student retain this ability, instead of being entirely overridden by the architecture-agnostic knowledge.

To this end, we design a decoupler $h(\cdot)$ that buffers the student's internal representation from arch-agnostic information extraction. It prevents the student embeddings from being entirely exposed to and dictated by the RSD objective.
Our architecture-agnostic knowledge decoupling (AAD) module is simple: it consists of two FC layers, $h_{e}(\cdot)$ and  $h_{a}(\cdot)$, joined by BatchNorm and GeLU activation. $h_{e}(\cdot)$ is an \textit{expander} that transforms the student embedding to a higher dimensional space.  $h_{a}(\cdot)$ is an \textit{adaptor} that aligns the expanded embedding to the dimension of the teacher embedding. Unlike OFA which uses different projection modules for different architectures, AAD is a \textit{one-size-fits-all module} and is \textit{more parameter-efficient} (see Figure~\ref{fig:efficiency}). 
After distillation, it is discarded and there is no additional overhead introduced at inference.

To summarise, the role of our AAD module is twofold: 1) It aligns the student and teacher embedding dimension to enable the calculation of the RSD loss. 2) It decouples the student internal representation from the one used by the RSD objective for extracting arch-invariant information in order to preserve student-exclusive knowledge.

\paragraph{Full objective.}
Putting all designs together, we define the redundancy suppression distillation (RSD) loss as: 
\begin{equation}
    \mathcal{L}_{\text{RSD}} = d(\mathbf{P}(h(\mathbf{z}^s), \mathbf{z}^t), \mathbf{T})
\end{equation}
for which we use the MSE for $d(\cdot)$. The full optimisation objective for our framework is simply a combination of the CE loss and our RSD loss:
\begin{equation}
    \mathcal{L} = \mathcal{L}_{\text{CE}} + \lambda \mathcal{L}_{\text{RSD}}
\end{equation}
where $\lambda$ is a balancing weight.
The RSD loss can be implemented in about 8 lines in PyTorch. A pseudocode snippet for it is provided in Algorithm~\ref{alg:rrd}.

\begin{table*}[t]
  \centering
  \small
  \renewcommand\tabcolsep{3.5pt}
  \renewcommand{\arraystretch}{0.99}
  \resizebox{0.79\textwidth}{!}{
  \begin{tabular}{cc cc cccc cccc>{\columncolor[gray]{0.85}}c}
    \toprule
    \multirow{2}{*}{Teacher} & \multirow{2}{*}{Student} & \multicolumn{2}{c}{From Scratch} & \multicolumn{4}{c}{Logits-based} & \multicolumn{5}{c}{Feature-based} \\
    \cmidrule(lr){3-4} \cmidrule(lr){5-8} \cmidrule(lr){9-13} &  & T.    & S.   & KD  & DKD & DIST & OFA & FitNets & CC & RKD & CRD & \textbf{RSD} \\
    \midrule
    \multicolumn{2}{c}{\emph{CNN-based students}} \\
    \midrule
    DeiT-T  & ResNet18 & 72.19 & 69.75 & 70.22 & 69.39 & 70.64  & \underline{71.34} & 70.44   & 69.77 & 69.47 & 69.25 & \textbf{71.70} \\ 
    Swin-T & ResNet18  & 81.35                               & 69.75                           & 71.14             & 71.10                          & 70.91              & \underline{71.85} & 71.18    & 70.07 & 68.89 & 69.09  & \textbf{72.13} \\ 
    Mixer-B/16  & ResNet18            & 76.58                                & 69.75                           & 70.89             & 69.89                          & 70.66                          & \underline{71.38} & 70.78  & 70.05 & 69.46 & 68.40  & \textbf{71.41} \\ 
    DeiT-T                   & MobileNetV2              & 72.19                            & 68.87                           & 70.87             & 70.14                          & 71.08              & \underline{71.39} & 70.95 & 70.69 & 69.72 & 69.60  & \textbf{72.18} \\ 
    Swin-T                   & MobileNetV2              & 81.35                               & 68.87                           & 72.05             & 71.71                          & 71.76                          & \underline{72.32} & 71.75 & 70.69 & 67.52 & 69.58  & \textbf{72.36} \\ 
    Mixer-B/16 & MobileNetV2 & 76.55                             & 68.87                           & \underline{71.92}             & 70.93                          & 71.74                          & \bf{72.12} & 71.59                            & 70.79 & 69.86 & 68.89  & 71.90 \\ 
    \toprule
    \multicolumn{2}{c}{\emph{Transformer-based students}}                                                                                      \\
    \midrule
    ResNet50 & DeiT-T               & 80.33                             & 72.17                           & 75.10             & 75.60\rlap{$^*$} & 75.13\rlap{$^*$} & \bf{76.55}\rlap{$^*$} & 75.84                & 72.56 & 72.06 & 68.53  & \underline{76.17}\rlap{$^*$} \thinspace \\ 
    ConvNeXt-T & DeiT-T & 82.05   & 72.17 & 74.00             & 73.95                          & 74.07              & \underline{74.41} & 70.45                            & 73.12 & 71.47 & 69.18  & \textbf{74.46} \\ 
    Mixer-B/16 & DeiT-T & 76.55    & 72.17 & 74.16 & 72.82 & 74.22  & \textbf{74.46} & \underline{74.38} & 72.82 & 72.24 & 68.23 & 74.26 \\ 
    ResNet50 & Swin-N & 80.33 & 75.53                           & 77.58             & 78.23\rlap{$^*$} & 77.95\rlap{$^*$} & \underline{78.64}\rlap{$^*$} & 78.33                & 76.05 & 75.90 & 73.90  & \textbf{78.78}\rlap{$^*$}\thinspace \\ 
    ConvNeXt-T  & Swin-N & 82.05  & 75.53  & 77.15 & 77.00  &  77.25 & \underline{77.50} & 74.81                       & 75.79 & 75.48 & 74.15  & \textbf{77.70} \\ 
    Mixer-B/16 & Swin-N & 76.55  & 75.53 & 76.26  & 75.03 & 76.54  & \textbf{76.63} & 76.17                         & 75.81 & 75.52 & 73.38  & \underline{76.55} \\ 
    \toprule
    \multicolumn{2}{c}{\emph{MLP-based students}}     \\ 
    \midrule
    ResNet50                 & ResMLP-S12               & 80.33                             & 76.65                           & 77.41             & 78.23\rlap{$^*$} & 77.71\rlap{$^*$} & \textbf{78.53}\rlap{$^*$} & 78.13                & 76.21 & 75.45 & 73.23  &\underline{78.32}\rlap{$^*$} \thinspace \\ 
    ConvNeXt-T & ResMLP-S12 & 82.05 & 76.65 & 76.84             & 77.23 & 77.24 & \underline{77.53} & 74.69              & 75.79 & 75.28 & 73.57  & \textbf{78.41} \\ 
    Swin-T & ResMLP-S12               & 81.35                             & 76.65                           & 76.67             & 76.99                          & 77.25            & \underline{77.31} & 76.48 & 76.15 & 75.10 & 73.40  & \textbf{77.61} \\ 
    \midrule
    \textit{Average Gain} & - & - & - & \textit{+1.55} & \textit{+1.29} & \textit{+1.68} & \textit{+2.20} & \textit{+1.14} & \textit{+0.49} & \textit{-0.37} & \textit{-1.77}  & \textbf{\textit{+2.34}} \\ 
    \bottomrule
  \end{tabular}} \vspace{-0.1cm}
  \caption{\textbf{Cross-architecture distillation results on ImageNet-1k.} Best results are in bold and second best results underlined. $*$ denotes results achieved by combining with FitNets~\cite{fitnets} following OFA~\cite{ofa}.}
  \label{tab:imagenet}
  \vspace{-0.1cm}
\end{table*}

\paragraph{Discussion.} We discuss why we choose the penultimate-layer embeddings for our RSD optimisation. The foremost reason is that we would like to avoid the whole complex operations involved when working with intermediate features, which is precisely what makes OFA complicated. 
As discussed in Sections~\ref{sec:intro} and~\ref{sec:prelim},
OFA had to introduce tailored operations to reconcile heterogeneous features into unified dimensionality, such as depth-wise separable convolution for CNN features, and attention blocks and token merging blocks (and other token operations) for ViT and MLP features. By contrast, the penultimate-layer representations are always 1-D and are neither feature-map-like nor token-like, saving us from all the architecture-specific operations and designs. In this sense, RSD is a \textit{more universal cross-architectural distillation method} than OFA. 
These embeddings are also smaller compared to intermediate features, which are resource efficient.
Obtained near the network output, they are also less arch-specific than features from earlier layers, which puts us in a better position to extract more arch-invariant information. 
\section{Experiments}

\begin{table*}[t]
\begin{minipage}[t]{0.32\textwidth}
\centering
\renewcommand{\arraystretch}{0.99} 
\resizebox{\textwidth}{!}{
\renewcommand\tabcolsep{6pt}
\begin{tabular}{ccc>{\columncolor[gray]{0.85}}c} \toprule
     \multirow{2}{*}{Design} & \multirow{2}{*}{\shortstack{Swin-T$\to$\\ResNet18}} & \multirow{2}{*}{\shortstack{ConvNeXt-T$\to$\\ResMLP-S12}} \vspace{-0.8mm} \\
     & & \\
    \midrule
    Baseline & 74.01 & 76.65 \\
    + RSD-corr & 80.65 & 83.40  \\ 
    + RSD-decorr & \textbf{83.92} & \textbf{84.21} \\
    \bottomrule
  \end{tabular}} \vspace{-0.6mm}
  \caption{\textbf{Effect of RSD.}}  \label{tab:abl_rsd}
\end{minipage}
\hfill
\begin{minipage}[t]{0.33\textwidth}
\centering
\renewcommand{\arraystretch}{1.0} 
\resizebox{0.98\textwidth}{!}{
\renewcommand\tabcolsep{7pt}
\begin{tabular}{ccc>{\columncolor[gray]{0.85}}c} \toprule
   \multirow{3}{*}{Design}   & CIFAR-100 & ImageNet-1k \vspace{0.6mm} \\  \cline{2-3} 
    &  \multirow{2}{*}{\shortstack{ViT-S$\to$\\ResMLP-S12}} & \multirow{2}{*}{\shortstack{ConvNeXt-T$\to$\\Mixer-B/16}} \vspace{-1.0mm} \\
    & & \\
    \midrule
    RSD & \textbf{82.94} & \textbf{80.73}  \vspace{-0.1mm} \\
    w/o AAD  & 82.26 & 79.93 \vspace{-0.5mm} \\
    \bottomrule
  \end{tabular}} \vspace{-0.2mm}
  \caption{\textbf{Effect of AAD.}} \label{tab:abl_aad}
\end{minipage} \vspace{-0.1cm}
\hfill
\begin{minipage}[t]{0.34\textwidth}
\centering
\renewcommand{\arraystretch}{0.98} 
\resizebox{\textwidth}{!}{
\renewcommand\tabcolsep{1.0pt}
  \begin{tabular}{cccc>{\columncolor[gray]{0.85}}c}
    \toprule
     \multirow{2}{*}{Logit loss} & \multirow{2}{*}{\shortstack{Swin-T$\to$\\ResNet18}} & \multirow{2}{*}{\shortstack{Mixer-B/16$\to$\\DeiT-T}} & \multirow{2}{*}{\shortstack{ConvNeXt-T$\to$\\ResMLP-S12}} \vspace{-1.0mm} \\
    & & & \\
    \midrule
    w/ KD~\cite{kd} & 78.74 & 71.36 & 72.25 \vspace{-0.3mm} \\
    w/ DKD~\cite{dkd} & 80.26 & 73.44 & 73.22 \vspace{-0.7mm} \\
    w/ OFA~\cite{ofa} & 80.60 & 70.69 & 78.87 \vspace{-0.7mm} \\ 
    \textbf{w/ RSD} & \textbf{83.23} & \textbf{77.22} & \textbf{81.15} \vspace{-0.6mm} \\
    \bottomrule
  \end{tabular}} \vspace{-0.7mm}
  \caption{\textbf{RSD as a strong logit distiller.}}
  \label{tab:rrd_logit}
\end{minipage} \vspace{-0.1cm}
\vspace{-0.0cm} \label{tab:ablation}
\end{table*}

\subsection{Experimental setup}

\paragraph{Implementations.}
We conduct cross-architecture distillation using different pairs of CNN, Transformer, and MLP models. We use ResNet~\cite{resnet}, MobileNet~\cite{mobilenetv2}, and ConvNeXt~\cite{convnext} for CNN models, ViT~\cite{vit}, Swin Transformer~\cite{swin}, and DeiT~\cite{deit} for Transformer, and ResMLP~\cite{resmlp} and MLP-Mixer~\cite{mlpmixer} for MLP. Experiments are conducted on the CIFAR-100~\cite{cifar100} and ImageNet-1k~\cite{imagenet} datasets, following the configurations of OFA~\cite{ofa}. More details are provided in the supplementary material. 

\paragraph{Baselines.} Under different setups, we compare our method against established generic KD methods (\textit{i.e.}, KD~\cite{kd}, DKD~\cite{dkd}, DIST~\cite{dist}, FitNets~\cite{fitnets}, CC~\cite{cckd}, RKD~\cite{rkd}, and CRD~\cite{crd}) and, notably, the cross-architectural pioneer OFA~\cite{ofa}.

\subsection{CIFAR-100 results}
Table~\ref{tab:cifar} presents the evaluation results for a total of 12 cross-architecture teacher-student pairs on CIFAR-100. It can be seen that the proposed method consistently outperforms all prior knowledge distillation methods, including OFA~\cite{ofa}.
It is noteworthy that over several distillation pairs, the margin RSD leads OFA by is several times that of OFA over prior arts. For example, RSD leads OFA by 3.38\% for Swin-T-to-ResNet18 distillation, where OFA is only 0.28\% higher than DKD~\cite{dkd}. 
For ViT-S-to-MobileNetV2 distillation, where OFA brings a mere 0.31\% improvement over CRD~\cite{crd}, RSD achieves an impressive 3.23\% advantage. 
The highest performance gain is observed on ConvNeXt-T-to-DeiT-T distillation, where RSD leads OFA by a surprising margin of 6.70\% -- nearly the gap between OFA and an un-distilled student. These results highlight the effectiveness of RSD, especially given that RSD is orthogonal to OFA and uses only a fraction of its parameters.

\subsection{ImageNet-1k results}
We evaluate our method on the large-scale ImageNet-1k dataset, using 15 heterogenous teacher-student pairs. As shown in Table~\ref{tab:imagenet}, RSD achieves state-of-the-art performance on a majority of the teacher-student pairs. Notably, RSD yields particularly large gains on certain pairs. For instance, when distilling from ConvNeXt-T to ResMLP-S12, RSD is 0.88\% higher than OFA, given that OFA only leads the earliest generic distillation method KD~\cite{kd} by 0.69\%. For ConvNeXt-T-to-Swin-N distillation, RSD has a 0.20\% advantage over OFA, when OFA is only 0.25\% higher than DIST~\cite{dist}, by using only 9\% as many extra parameters as OFA.
RSD achieves an average gain of 2.34\% across the 15 heterogeneous model pairs, highest amongst all methods.

\begin{figure}[t] \centering \vspace{-2mm}
\includegraphics[width=\linewidth]{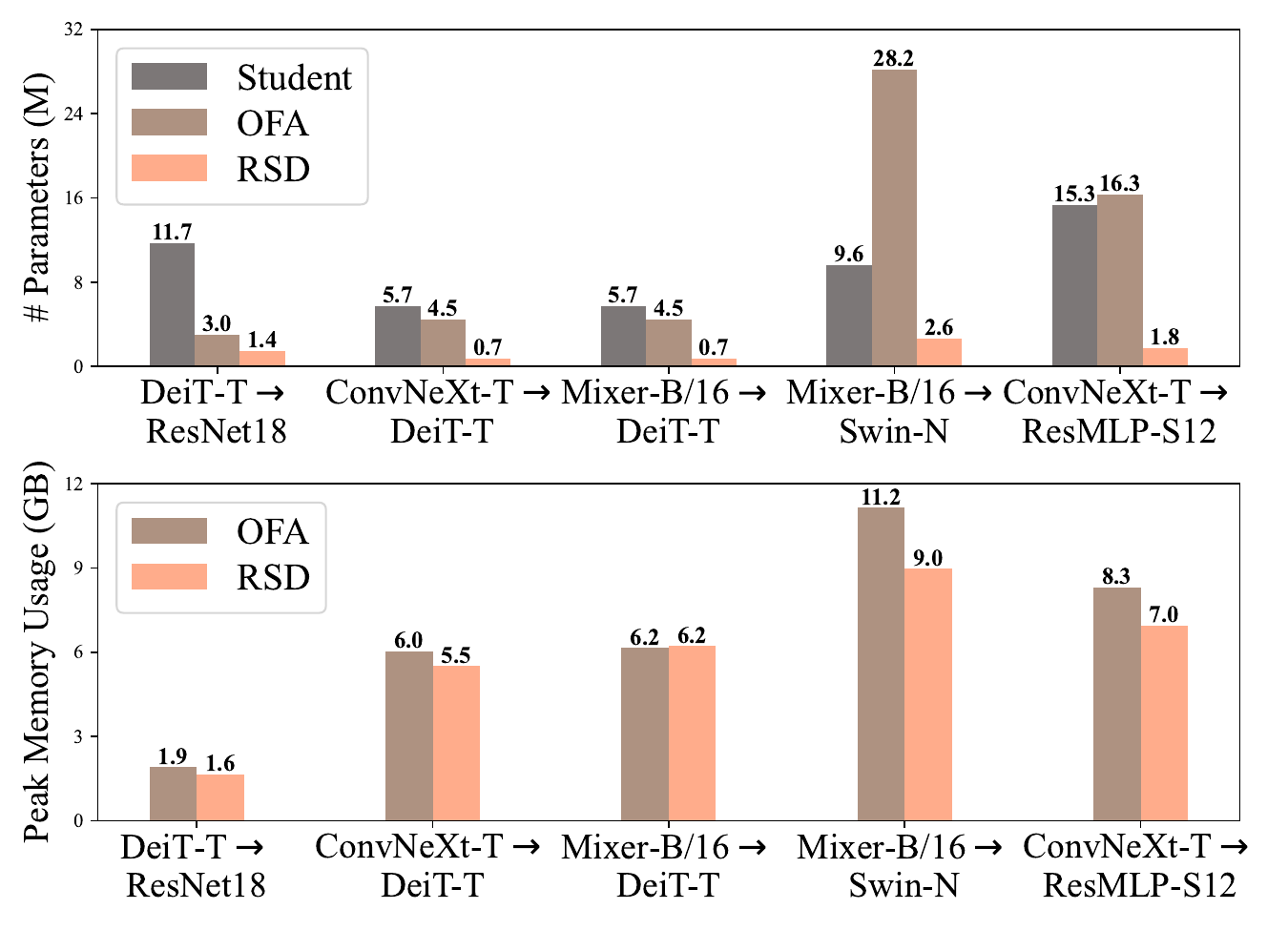} \vspace{-6.6mm}
\caption{\textbf{{A comparison of the computational overheads.}}} 
\label{fig:efficiency}
\end{figure}

\subsection{Further analysis}

\paragraph{Effect of proposed designs.}
We conduct ablation experiments to gauge the individual effectiveness of the proposed components in Tables~\ref{tab:abl_rsd} and~\ref{tab:abl_aad}. 
It is noteworthy when there is a dimensionality mismatch between the teacher and student embeddings, it is impossible to ablate the AAD block to assess the sole effect of the RSD objective. To enable such analysis, we set up teacher-student pairs with identical penultimate embedding dimension, such that the unadapted student representation may get directly exposed to the architecture-agnostic knowledge. The ablation results demonstrate the effectiveness of the core designs of redundancy suppression distillation and architecture-agnostic knowledge decoupling. 

\paragraph{RSD as a strong logit distiller.} While we have chosen to employ RSD between the informative feature embeddings from the teacher and the student,  we further discovered that RSD is also a strong logit distiller.  
We conducted experiments to directly apply the proposed RSD loss over the logits. For head-to-head comparisons, we also apply the logit distillation losses from KD, DKD, and OFA independently. For OFA, we adopt the set of losses computed from projected logits to ablate any impact of the logit KD loss of ~\cite{kd}. From Table~\ref{tab:rrd_logit}, RSD's logit variant itself (without AAD or any projectors) is a strong logit distillation objective. It outperforms individual existing logit distillation objectives with considerable margins, particularly OFA. This also reveals that OFA's performance heavily relies on the logit KD loss (\textit{i.e.}, 3rd term in Equation~\ref{eq:ofa}). By contrast, \textit{RSD is free from such constraint and does not rely on an additional logit matching loss}.
Noticeably, this one single RSD loss still surpasses the full formulations of OFA in Table~\ref{tab:cifar} that encompass costly and tailored projector branches, adaptive target knowledge enhancer, and the KD loss.

\begin{table}[t]
  \centering
  \renewcommand\tabcolsep{5pt}
  \renewcommand{\arraystretch}{1.0}
  \resizebox{0.38\textwidth}{!}{
  \begin{tabular}{cccc>{\columncolor[gray]{0.85}}c}
    \toprule
    \multirow{3}{*}{Logit loss} & \multicolumn{2}{c}{CIFAR-100} & ImageNet-1k \vspace{0.6mm} \\  \cline{2-4} \vspace{1mm}
     & \multirow{2}{*}{\shortstack{Swin-T$\to$\\ResNet18}} & \multirow{2}{*}{\shortstack{ConvNeXt-T$\to$\\ResMLP-S12}} & \multirow{2}{*}{\shortstack{Deit-T$\to$\\ResNet18}} \vspace{-2mm} \\
    & & & \\
    \midrule
    OFA & 80.54 & 81.22 & 70.64\textsuperscript{†} \\
    \textbf{OFA+RSD} & \textbf{83.35} & \textbf{84.46} & \textbf{71.21} \\
    \bottomrule
  \end{tabular}} \vspace{-0.5mm}
  \caption{\textbf{Integrating the RSD objective into OFA.}}
  \label{tab:rrd_ofa}
\end{table}

\begin{table}[t]
  \centering \vspace{1mm}
  \renewcommand{\arraystretch}{0.92}
   \resizebox{0.27\textwidth}{!}{
  \begin{tabular}{ccc}
    \toprule
    \multirow{2}{*}{Stage}  & \multirow{2}{*}{\shortstack{Swin-T$\to$\\ResNet18}}   & \multirow{2}{*}{\shortstack{Mixer-B/16$\to$\\DeiT-T}} \\
    & & \\
    \midrule
    $\varnothing$  & 83.92 & 76.45 \\
    \{4\}          & 84.39 & 76.83 \\
    \{3,4\}        & 84.81 & 76.37 \\
    \{2,3,4\}      & 84.63 & \textbf{77.17} \\
    \{1,2,3,4\}    & \textbf{85.45} & 76.90 \\
    \bottomrule
  \end{tabular}} \vspace{-0.0cm}
  \caption{\textbf{Integrating OFA into RSD at different network stages.}} \label{tab:rrd_stage} \vspace{-1mm}
\end{table}

\paragraph{Compatibility with OFA.} We investigate whether our spirit of redundancy suppression is complementary to the method of OFA. To make RSD integrated with OFA, we simply replace all OFA losses in it with our RSD loss. This involves using the proposed RSD loss to replace OFA's KD loss at logit level and the OFA losses at each network stage. Effectively, RSD now reduces the redundancy between full-stage intermediate features of the teacher and the student, using OFA's projected logits as a proxy.
The results are provided in Table~\ref{tab:rrd_ofa}, where † denotes the result reproduced by ourselves for a fair analysis. We find that the RSD objective can be seamlessly integrated with OFA and leads to substantial performance gains. In addition, we perform another set of experiments to extrapolate RSD to OFA's multi-stage features.
In Table~\ref{tab:rrd_stage}, we start with our RSD baseline where no intermediate features are utilised, denoted by $\varnothing$ (\textit{i.e.}, only penultimate-layer embeddings are used). We then gradually incorporate intermediate features from various network stages, and apply our RSD objective over OFA-projected pseudo logits. As can be seen, RSD is compatible and generally benefited by additional intermediate features.

\begin{table}[t]
  \centering
  \renewcommand{\arraystretch}{0.88}
   \resizebox{0.34\textwidth}{!}{
  \begin{tabular}{cccc>{\columncolor[gray]{0.85}}c}
    \toprule
   & \shortstack{T.: ResNet34\\S.: ResNet18} & \shortstack{T.: ResNet50\\S.: MobileNetV1} \\ \midrule
    KD~\cite{kd} & 70.66 & 68.58 \\
    OFD~\cite{ofd}  & 70.81 & 71.25 \\
    CRD~\cite{crd}  & 71.17 & 71.37 \\
    RKD~\cite{rkd}  & 71.34 & 71.32 \\
    CAT-KD~\cite{catkd} & 71.26 & 72.24 \\
    SimKD~\cite{simkd} & 71.59 & 72.25 \\
    ReviewKD~\cite{reviewkd} & 71.61 & 72.56 \\
    DKD~\cite{dkd} & 71.70 & 72.05 \\
    SDD~\cite{sdd} & 71.14 & 72.24 \\
    DIST~\cite{dist}  & 72.07 & \textbf{73.24} \\
    OFA~\cite{ofa}  & \underline{72.10} & - \\
    RSD  & \textbf{72.18} & \underline{73.08} \\
    \bottomrule
  \end{tabular}} \vspace{-1.5mm}
  \caption{\textbf{Same-architecture distillation results on ImageNet-1K.} The best result is in bold and second best underlined.}
  \label{tab:homo}
  \vspace{-0mm}
\end{table}

\paragraph{Computational cost.}
We compare the computational costs of our method to OFA in terms of the number of extra parameters introduced and peak GPU memory usage on ImageNet-1k. The measurements are visualised in Figure~\ref{fig:efficiency} for multiple teacher-student pairs. Where OFA introduces an amount of extra network parameters on par with or more than that of the student itself, \textit{our method saves up to $10\times$ in parameter count to achieve better performance}. In terms of peak GPU memory usage, RSD is also better off on average. RSD's overhead benefit persists on the CIFAR-100 dataset, yet with even larger performance gains.

\paragraph{Same-Architecture Evaluation.}
Following OFA, we also perform evaluation under the same-architecture distillation set-up. We additionally compare with established generic KD methods OFD~\cite{ofd}, ReviewKD~\cite{reviewkd}, CAT-KD~\cite{catkd}, SimKD~\cite{simkd}, and the recent SDD~\cite{sdd}.
As shown in Table~\ref{tab:homo}, our method delivers competitive or even better results compared to OFA and leading same-architecture distillation methods, including SDD, DKD, and DIST. We notice that the advantage of RSD is more prominent on cross-architecture distillation. This observation reinforces our initial motivation, where our designs are rooted in theories for learning architecture-invariant knowledge from heterogeneous representations.
Overall, the same-architecture distillation results highlight the significance of our findings to a broader knowledge distillation community.

\begin{figure}[t] \centering
\begin{subfigure}[b]{\linewidth}
\includegraphics[width=\linewidth, trim=2 2 2 2,clip]{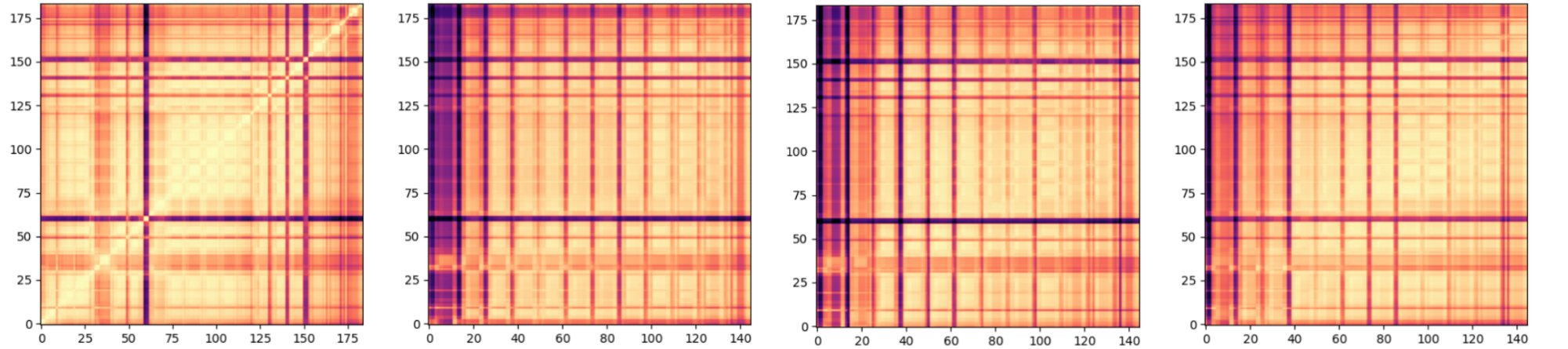}
\end{subfigure}
\begin{subfigure}[b]{\linewidth}
\includegraphics[width=\linewidth, trim=2 2 2 2,clip]{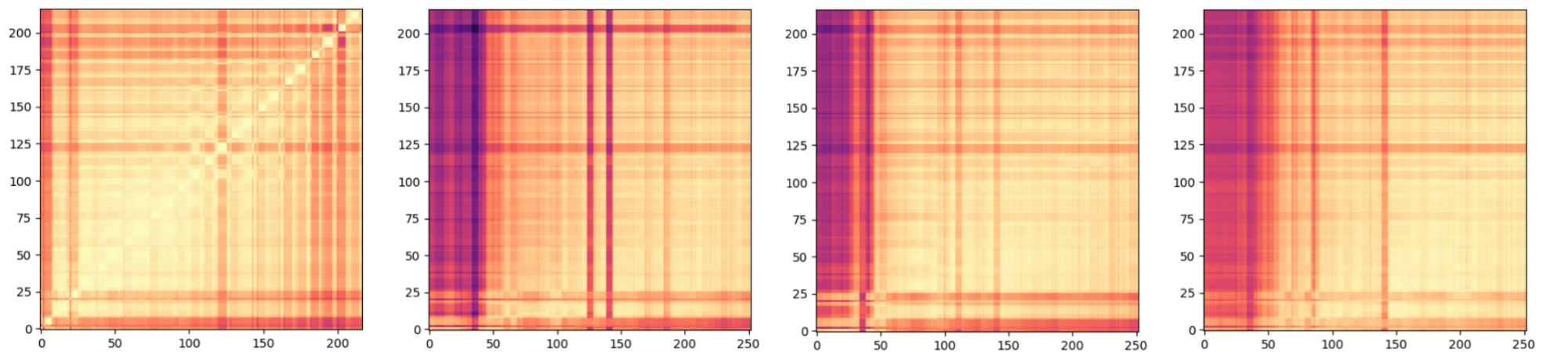}
\end{subfigure}
\vspace{-6.5mm}
\caption*{\scriptsize \qquad  T. v.s. T. \qquad \enspace  T. v.s. S. (w/o KD) \quad \enspace T. v.s. S. (OFA) \qquad T. v.s. S. (RSD)} \vspace{-1mm}
\caption{\textbf{Cross-architectural feature similarities measured by CKA.} Brighter colours indicate higher similarity and vice versa. Top: ConvNeXt-T teacher and ResMLP-S12 student. Bottom: ViT-S teacher and MobileNetV2 student.} 
\label{fig:cka} \vspace{-0mm}
\end{figure}

\paragraph{Visualisation of cross-architectural feature similarity.} We employ centered kernel alignment (CKA)~\cite{cka} to assess the effectiveness of the proposed method. CKA is a similarity metric that can accommodate inputs of different dimensions. It enables analysis of cross-architectural feature similarity between different layers and stages in heterogeneous network pairs.
Figure~\ref{fig:cka} visualises the CKA scores between layers of different teacher-student pairs. We observe that under the first set-up, RSD significantly increases the representation similarity between heterogeneous architectures at various stages.
Whereas under the second set-up, some shallow feature dissimilarities in OFA are slightly alleviated, while some others are enlarged. This is justified because unlike OFA, our method does not directly access or process those intermediate features. Overall, RSD notably increases feature similarity at middle and deep layers. 

\paragraph{Limitations \& future work.} Despite its simple formulation, RSD's performance can sometimes be sensitive to hyperparameters $\lambda$ and $\kappa$. Compared to CIFAR-100, its advantages are less prominent on larger-scale datasets such as ImageNet-1k, which is in part due to the unused 2-D features. RSD in its current design, albeit efficient and universal, only utilises the 1-D embeddings. While this offers benefits in efficiency and design, it fails to harness the rich spatial context associated with the 2-D feature maps, and is therefore unable to be directly extended to tasks that heavily demand spatial information such as object detection. We leave the extension to such tasks for future investigation.
\section{Conclusion}
We introduced RSD, a simple approach for cross-architecture knowledge distillation based on redundant knowledge suppression. RSD employs invariance maximisation and feature decorrelation objectives to extract arch-agnostic knowledge  common to heterogeneous architectures. It also allows student-exclusive patterns to be  retained rather than entirely overridden by RSD through a lightweight decoupling module. RSD achieves superior performance over the recent cross-architectural baseline of OFA with a fraction of its parameter overhead, while avoiding architecture-tailored operations in OFA.

\newpage

\paragraph{Acknowledgement.} This work was supported in part by NSFC (62322113, 62376156), Shanghai Municipal Science and Technology Major Project (2021SHZDZX0102), and the Fundamental Research Funds for the Central Universities.

{
\small \bibliographystyle{ieeenat_fullname}
\bibliography{main}
}

\end{document}